\documentclass[11pt,a4paper]{article}
\usepackage{times,latexsym}
\usepackage{url}
\usepackage{enumitem}
\usepackage[T1]{fontenc}
\usepackage{subcaption}
\usepackage[nohyperref,acceptedWithA]{tacl2018v2}
\usepackage{graphicx}
\usepackage{xspace,mfirstuc,tabulary}
\usepackage{booktabs}
\usepackage{multirow}

\newif\iftaclinstructions
\taclinstructionsfalse
\iftaclinstructions
\renewcommand{\confidential}{}
\renewcommand{\anonsubtext}{(No author info supplied here, for consistency with
TACL-submission anonymization requirements)}
\newcommand{\instr}
\fi

\iftaclpubformat 

\else

\fi

\newcommand{\Sec}[1]{{Section \ref{section:#1}}}
\newcommand{\Eq}[1]{{Equation (\ref{eq:#1})}}
\newcommand{\moses}{Moses}
\newcommand{\marian}{Marian}

\title{Unsupervised Neural Machine Translation \\Initialized by Unsupervised Statistical Machine Translation}
\author{Benjamin Marie \qquad Atsushi Fujita \\
	National Institute of Information and Communications Technology \\
	3-5 Hikaridai, Seika-cho, Soraku-gun, Kyoto, 619-0289, Japan\\
	\textsf{\{bmarie, atsushi.fujita\}@nict.go.jp}
\\}

\date{}

\begin{document}
\maketitle
\begin{abstract}
Recent work achieved remarkable results in training neural machine translation (NMT) systems in a fully unsupervised way, with new and dedicated architectures that rely on monolingual corpora only. In this work, we propose to define unsupervised NMT (UNMT) as NMT trained with the supervision of synthetic bilingual data. Our approach straightforwardly enables the use of state-of-the-art architectures proposed for supervised NMT by replacing human-made bilingual data with synthetic bilingual data for training. We propose to initialize the training of UNMT with synthetic bilingual data generated by unsupervised statistical machine translation (USMT). The UNMT system is then incrementally improved using back-translation. Our preliminary experiments show that our approach achieves a new state-of-the-art for unsupervised machine translation on the WMT16 German--English news translation task, for both translation directions.
\end{abstract}

\section{Introduction}
Machine translation (MT) systems usually require a large amount of bilingual data, produced by humans, as supervision for training. However, finding such data remains challenging for most language pairs, as it may not exist or may be too costly to manually produce.

In contrast, a large amount of monolingual data can be easily collected for many languages, for instance from the Web.\footnote{See for instance the Common Crawl project: \url{http://commoncrawl.org/}} Previous work proposed many ways for taking advantage of the monolingual data in order to improve translation models trained on bilingual data.  These methods usually exploit existing accurate translation models and have shown to be useful especially when targeting low-resource language pairs and domains. However, they usually fail when the available bilingual data is too noisy or too small to train useful translation models. In such scenarios, the use of pivot languages or unsupervised machine translation are possible alternatives.

Recent work has shown remarkable results in training MT systems using only monolingual data in the source and target languages.  Unsupervised statistical (USMT) and neural (UNMT) machine translation have been proposed \citep{artetxe2018unsupervised,DBLP:journals/corr/abs-1804-07755}. 
State-of-the-art USMT \citep{artetxe2018unsupervised,DBLP:journals/corr/abs-1804-07755} uses a phrase table induced from source and target phrases, extracted from the monolingual data, paired and scored using bilingual word, or $n$-gram, embeddings trained without supervision. This phrase table is plugged in a standard phrase-based SMT framework that is used to translate target monolingual data into the source language, i.e., performing a so-called back-translation. The translated target sentences and their translations in the source language are paired to form synthetic parallel data and to train a source-to-target USMT system.  This back-translation/re-training step is repeated for several iterations to refine the translation model of the system.\footnote{Previous work did not address the issue of convergence and rather fixed the number of iterations to perform for these refinement steps.}
On the other hand, state-of-the-art UNMT \citep{DBLP:journals/corr/abs-1804-07755} uses bilingual sub-word embeddings. They are trained on the concatenation of source and target monolingual data in which tokens have been segmented into sub-word units using, for instance, byte-pair-encoding (BPE) \citep{sennrich-haddow-birch:2016:P16-12}. This method can learn bilingual embeddings if the source and target languages have in common some sub-word units. The sub-word embeddings are then used to initialize the lookup tables in the encoder and decoder of the UNMT system. Following this initialization step, UNMT mainly relies on denoising autoencoder as language model during training and on latent representation shared across the source and target languages for the encoder and the decoder.

While the primary target of USMT and UNMT is low-resource language pairs, their possible applications for these language pairs remain challenging, especially for distant languages,\footnote{Mainly due to the difficulty of training accurate unsupervised bilingual word/sub-word embeddings for distant languages \citep{P18-1072}.} and have yet to be demonstrated. On the other hand, unsupervised MT achieves impressive results on resource-rich language pairs, with recent and quick progresses, suggesting that it may become competitive, or more likely complementary, to supervised MT in the near future.

In this preliminary work, we propose a new approach for unsupervised MT to further reduce the gap between supervised and unsupervised MT. Our approach exploits a new framework in which UNMT is bootstrapped by USMT and uses only synthetic parallel data as supervision for training. The main outcomes of our work are as follows:
\begin{itemize}\itemsep=0mm
\item We propose a simplified USMT framework. It is easier to set up and train. We also show that using back-translation to train USMT is not suitable and underperform.
\item We propose to use supervised NMT framework for the unsupervised NMT scenarios by simply replacing true parallel data with synthetic parallel data generated by USMT. This strategy enables the use of well-established NMT architectures with all their features, without assuming any relatedness between source and target languages in contrast to previous work.
\item We empirically show that our framework leads to significantly better UNMT than USMT on the WMT16 German--English news translation task, for both translation directions.
\end{itemize}

\section{What is truly unsupervised in this paper?}
Since the term ``unsupervised'' may be misleading, we present in this section what aspects of this work are truly unsupervised.

As previous work, we define ``unsupervised MT'' as MT that does not use human-made translation pairs as bilingual data for training. Nonetheless, MT still needs some supervision for training. Our approach uses as supervision synthetic bilingual data generated from monolingual data. 

``Unsupervised'' qualifies only the training of MT systems on bilingual parallel data of which at least one side is synthetic. For tuning, it is arguably unsupervised in some of our experiments or supervised using a small set of human-made bilingual sentence pairs. We discuss ``unsupervised tuning'' in Section \ref{section:tuning}. For evaluation, it is fully supervised, as in previous work, since we use a human-made test set to evaluate the translation quality.

Even if our systems are trained without human-made bilingual data, we can still argue that the monolingual corpora used to generate synthetic parallel data have been produced by humans. Source and target monolingual corpora in our experiments (see \Sec{expset}) could include some comparable parts. Moreover, we cannot ensure that they do not contain any human-made translations from which our systems can take advantage during training. Finally, we use SMT and NMT architectures, set and use their hyper-parameters (for instance, the default parameters of the Transformer model) in our framework that have already shown to give good results in supervised MT.

\section{Simplified USMT}
\begin{figure}[t]
    \centering
            \includegraphics[width=0.9\linewidth]{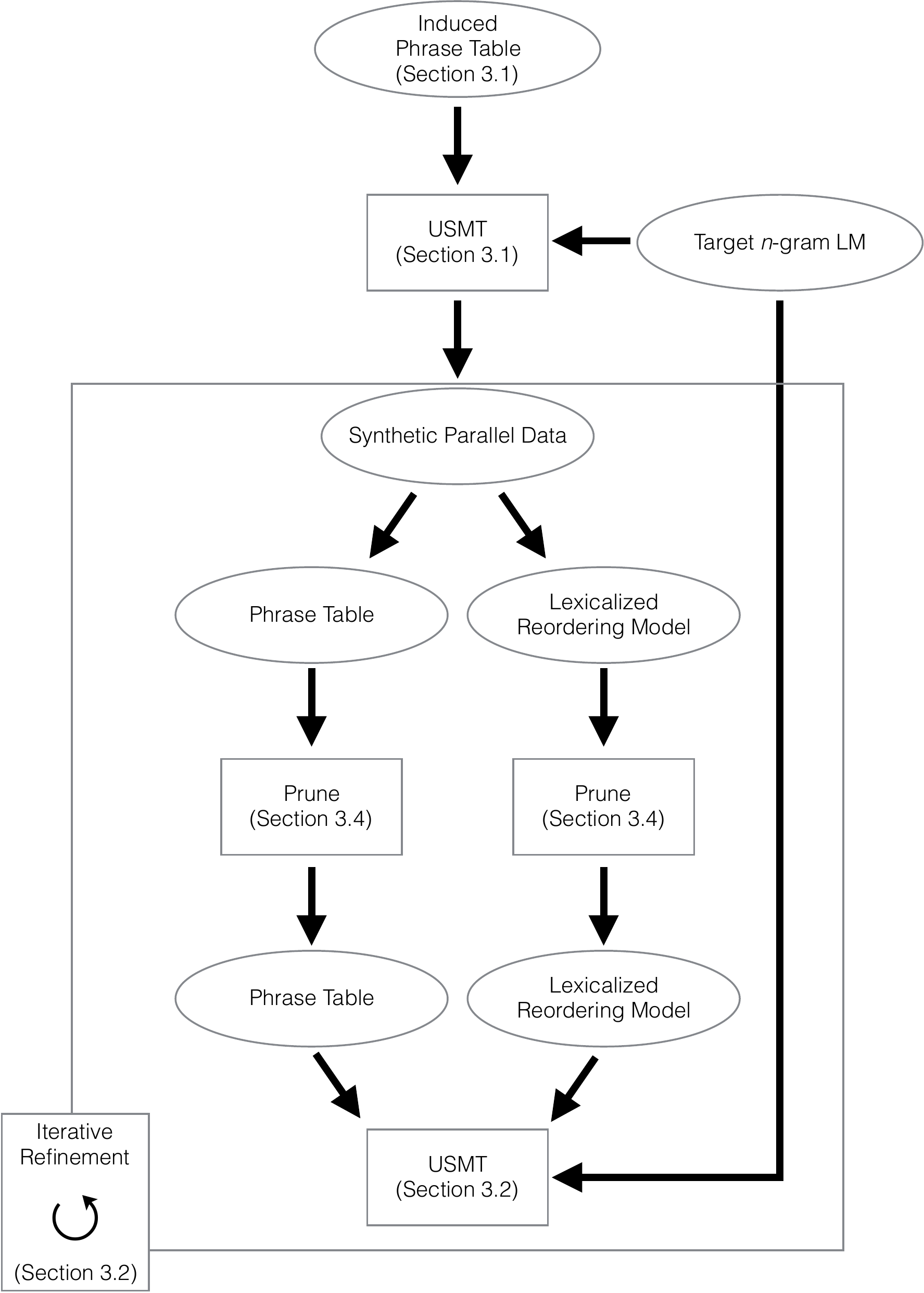}
    \caption{\label{fig:usmt} Our USMT framework.}
\end{figure}

Our USMT framework is based on the same architecture proposed by previous work \citep{artetxe2018unsupervised,DBLP:journals/corr/abs-1804-07755}: a phrase table is induced from monolingual data and used to compose the initial USMT system that is then refined iteratively using synthetic parallel data. We propose the following improvements and discussions to simplify the framework and make it faster with lighter models (see also Figure~\ref{fig:usmt}):
\begin{itemize}\itemsep=0mm
\item Section \ref{section:ipt}: we propose several modifications to rely more on compositional phrases and to simplify the phrase table induction compared to the method proposed by \citet{artetxe2018unsupervised}
\item Section \ref{section:tuning}: we discuss the feasibility of unsupervised tuning.
\item Section \ref{section:refinement}: we propose to replace the back-translation in the refinement steps with forward translation to improve translation quality and to remove the need of simultaneously training models for both translation directions.
\item Section \ref{section:pruning}: we propose to prune the phrase table to speed up the generation of synthetic parallel data during the refinement steps.
\end{itemize}

\subsection{Phrase table induction}
\label{section:ipt}

\begin{figure}[t]
    \centering
            \includegraphics[width=\linewidth]{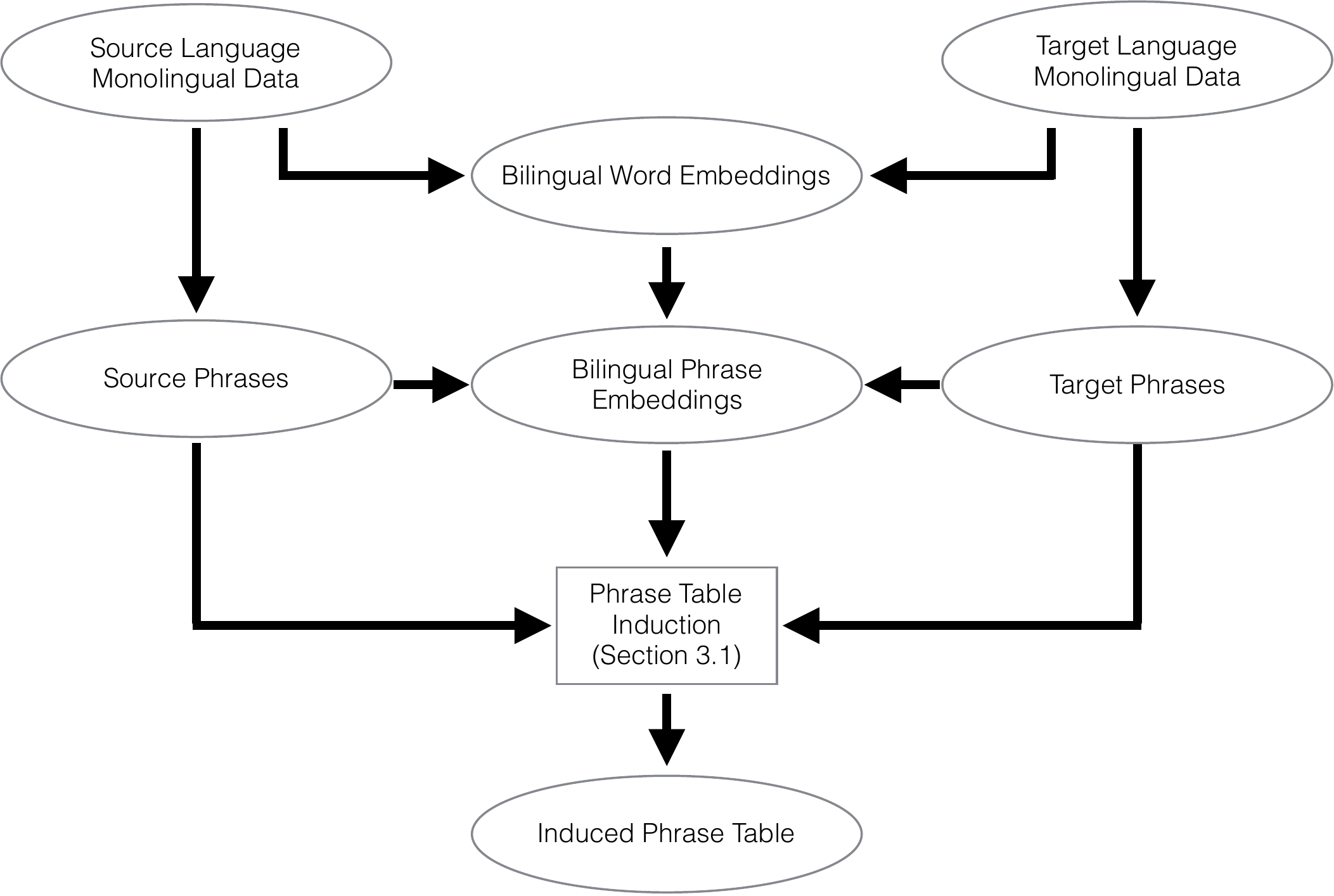}
    \caption{\label{fig:ipt} Phrase table induction.}
\end{figure}

As proposed by \citet{artetxe2018unsupervised} and \citet{DBLP:journals/corr/abs-1804-07755}, the first step of our approach for USMT is an unsupervised phrase table induction that only takes as inputs a set of source phrases, a set of target phrases, and their respective embeddings, as illustrated by Figure \ref{fig:ipt}.
\citet{artetxe2018unsupervised} regarded the most frequent unigrams, bigrams, and trigrams in the monolingual data as phrases. The embedding of each n-gram is computed with a generalization of the skipgram algorithm \citep{Mikolov:2013:DRW:2999792.2999959}. Then, source and target n-gram embedding spaces are aligned in the same bilingual embedding space without supervision \citep{P18-1073}. \citet{DBLP:journals/corr/abs-1804-07755}'s method also works at n-gram level, but computes phrase embeddings as proposed by \citet{N15-1176}: performing the element-wise addition of the embeddings of the component words of the phrase, also trained on the monolingual data and aligned in the same bilingual embedding space. This method can estimate embedding for compositional phrases but not for non-compositional phrases unlike \citet{artetxe2018unsupervised}'s method. Interestingly, \citet{artetxe2018unsupervised}'s method yields significantly better results at the first iteration of USMT, that uses the induced phrase table, but performs similarly to \citet{DBLP:journals/corr/abs-1804-07755}'s method after several refinement steps (see Section \ref{section:refinement}).

We choose to build USMT with an alternative method for phrase table induction. We adopt the method proposed by \citet{Marie:2018:PTI:3184403.3168054}, except that we remove the supervision using a bilingual word lexicon. First, phrases are collected using the following equation \citep{Mikolov:2013:DRW:2999792.2999959}:
\begin{equation}
\label{eq:phrase}
\mathit{score}(w_i w_j) = \frac{\mathit{freq}(w_i w_j)-\delta}{\mathit{freq}(w_i) \times \mathit{freq}(w_j)},
\end{equation}
where $w_i$ and $w_j$ are two consecutive tokens or phrases in the monolingual data, $\mathit{freq}(\cdot)$ the frequency of the given token or phrase, and $\delta$ a discounting coefficient for preventing the retrieval of phrases composed of very infrequent tokens. Consecutive tokens/phrases having a higher score than a pre-defined threshold are regarded as new phrases,\footnote{This transformation is performed by simply replacing the space between the two tokens/phrases with an underscore.} and a new pass is performed to obtain longer phrases.
The iteration results in the collection of much longer and meaningful phrases, i.e., not only very frequent sequences of grammatical words, rather than only short n-grams. In our experiments, we perform 6 iterations to collect phrases of up to 6 tokens.\footnote{We chose a maximum phrase length of 6, since this value is usually used as the maximum length in most state-of-the-art SMT frameworks.} \Eq{phrase} was originally proposed to identify non-compositional phrases. However, we choose to enforce the collection of more compositional phrases with a low $\delta$\footnote{We set $\delta=10$ in all our experiments.} for the following reasons:
\begin{itemize}\itemsep=0mm
\item very few phrases are actually non-compositional in standard SMT systems \citep{D12-1089},
\item most of them are not very frequent, and
\item useful representation of compositional phrases can easily be obtained compositionally \citep{N15-1176}.
\end{itemize}

To obtain the pairs of source and target phrases that populate the induced phrase table, we used the Equation proposed by \citet{DBLP:journals/corr/abs-1804-07755}:\footnote{We could not obtain results similar to the results reported in \citet{DBLP:journals/corr/abs-1804-07755} (the second version of their arXiv paper) by using their Equation (3) with $\beta=30$ as they proposed.  We have confirmed through personal communications with the authors that \Eq{prob}, as we wrote, with $\beta=30$, generates the expected results. We did not use the Equation computing $\phi$ in \citet{artetxe2018unsupervised}, since it produces negative value as a probability when cosine similarity is negative.}
\begin{equation}
\label{eq:prob}
p(t_j|s_i) = \frac{\exp{(\beta\cos(emb(t_j),emb(s_i))})}{\sum_k{\exp{(\beta\cos(emb(t_k),emb(s_i))})}} ,
\end{equation}
where $t_j$ is the $j$-th phrase in the target phrase list
and $s_i$ the $i$-th phrase in the source phrase list, $\beta$ a parameter to tune the peakiness of the distribution\footnote{We set $\beta=30$ since it is the default value proposed in the code released by \citet{DBLP:journals/corr/SmithTHH17}: \url{https://github.com/Babylonpartners/fastText_multilingual}} \citep{DBLP:journals/corr/SmithTHH17}, and $emb(\cdot)$ a function returning the bilingual embedding of a given phrase.

In this work, for a reasonably fast computation, we retained only the 300k most frequent phrases in each language and retained for each of them the 300-best target phrases according to \Eq{prob}.

Standard phrase-based SMT uses the following four translation probabilities for each phrase pair.
\begin{enumerate}[label=(\alph*)]\itemsep=0mm
\item $p(t_j|s_i)$: forward phrase translation probability
\item $p(s_i|t_j)$: backward phrase translation probability
\item $lex(t_j|s_i)$: forward lexical translation probability
\item $lex(s_i|t_j)$: backward lexical translation probability
\end{enumerate} 
These probabilities, except (a), need to be computed only for the 300-best target phrases for each source phrase that are already determined using (a). (b) is given by switching $s_i$ and $t_j$ in \Eq{prob}. To compute lexical translation probabilities, (c) and (d), given the significant filtering of candidate target phrases, we can adopt a more costly but better similarity score.
In this work, we compute them using word embeddings as proposed by \citet{N15-1138}:
\begin{equation}
\label{eq:problex}
lex(t_j|s_i) = \frac{1}{L}\prod_{l=1}^{L}\max_{k=1}^{K}p(t_j^k|s_i^l)
\end{equation}
where $K$ and $L$ are the number of words in $t_j$ and $s_i$, respectively, and $p(t_j^k|s_i^l)$ the translation probability of the $k$-th target word $t_j^k$ of $t_j$ given the $l$-th source word $s_i^l$ of $s_i$ given by \Eq{prob}.  This phrase-level lexical translation probability is computed for both translation directions.
Note that, unlike \citet{N15-1138} and \citet{C16-1109}, we do not use a threshold value under which $p(t_j^k|s_i^l)$ is ignored, since it would require some supervised fine-tuning to be set according to the translation task. In practice, even without this threshold value, our preliminary experiments showed significant improvements of translation quality by incorporating $lex(t_j|s_i)$ and  $lex(s_i|t_j)$ into the induced phrase table.

After the computation of the above four scores for each phrase pair in the induced phrase table, the phrase table is plugged in an SMT system to perform what we denote in the remainder of this paper as iteration 0 of USMT.

Computing lexicalized reordering models for the phrase pairs in the induced phrase table from monolingual data is feasible and helpful as shown by \citet{klementiev-EtAl:2012:EACL2012}. However, for the sake of simplicity, we do not compute these lexical reordering models for iteration 0.

\subsection{Discussion about unsupervised tuning}
\label{section:tuning}
State-of-the-art supervised SMT performs the weighted log-linear combination of different models \citep{och-ney:2002:ACL}. The model weights are tuned given a small development set of bilingual sentence pairs. 
For completely unsupervised SMT, we cannot assume the availability of this development set. In other words, model weights must be tuned without the supervision of manually produced bilingual data.

\citet{DBLP:journals/corr/abs-1804-07755} used some pre-existing default weights that work reasonably well. On the other hand, \citet{artetxe2018unsupervised} obtained better results by using 10k monolingual sentences paired with their back-translations as a development set. Nonetheless, to create this development set, they also relied on the same pre-exisintg default weights used by \citet{DBLP:journals/corr/abs-1804-07755}. To be precise, both used the default weights of the {\moses} framework \citep{P07-2045}.
In this preliminary work, we present results with supervised tuning and with the {\moses}'s default weights.

However, regarding the use of default weights as ``unsupervised tuning'' is arguable, since these default weights have been determined manually to work well for European languages. For translation between much more distant languages,\footnote{For instance, \citet{DBLP:journals/corr/abs-1804-07755} presented for Urdu--English only the results with supervised tuning.} these default weights would likely result in a very poor translation quality. We argue that unsupervised tuning remains one of the main issues in current approaches for USMT. 

Note that while creating large training bilingual data manually for a particular language pairs is very costly, which is one of the fundamental motivations of unsupervised MT, we can assume that a small set of sentence pairs required for tuning can be created at a reasonable cost.

\subsection{Refinement without back-translation}
\label{section:refinement}
\citet{artetxe2018unsupervised} and \citet{DBLP:journals/corr/abs-1804-07755} presented the same idea of performing so-called refinement steps. Those steps use USMT to generate synthetic parallel data to train a new phrase table, with refined translation probabilities. This can be repeated for several iterations to improve USMT. The initial system at iteration 0 uses the induced phrase table (see Section \ref{section:ipt}), while the following iterations use only a phrase table and a lexicalized reordering model trained on the synthetic parallel data generated by USMT. They both fixed the number of iterations.

\citet{artetxe2018unsupervised} and \citet{DBLP:journals/corr/abs-1804-07755} generated the synthetic parallel data through back-translation: a target-to-source USMT system was used to back-translate sentences in the target language, then the pairs of each sentence in the target language and its USMT output in the source language were used as synthetic parallel data to train a new source-to-target USMT system. This way of using back-translation has originally been proposed to improve NMT systems \citep{sennrich-haddow-birch:2016:P16-11} with a specific motivation to enhance the decoder by exploiting fluent sentences in the target language. In contrast, however, using back-translation for USMT lacks motivation. Since the source side of the synthetic parallel data, i.e., decoded results of USMT, is not fluent, USMT will learn a phrase table with many ungrammatical source phrases, or foreign words, that will never be seen in the source language, meaning that many phrase pairs in the phrase table will never be used. Moreover, possible and frequent source phrases, or even source words, may not be generated via back-translation and will be consequently absent from the trained phrase table.

We rather consider that the language model already trained on a large monolingual corpus in the target language can play a much more important role in generating more fluent translations.
This motivates us to perform the refinement steps on synthetic parallel data made of source sentences translated into the target language by the source-to-target system, i.e., ``forward translation,'' as opposed to back-translation. In fact, the idea of retraining an SMT system on synthetic parallel data generated by a source-to-target system has already been proven beneficial \citep{P07-1004}. 

At each iteration, we randomly sample new $N$ source sentences from the monolingual corpus and translate them with the latest USMT system to generate synthetic parallel data.

\subsection{Phrase table pruning}
\label{section:pruning}

Generating synthetic parallel data through decoding millions of sentences is one of the most computationally expensive parts of the refinement steps, requiring also a large memory to store the whole phrase table.\footnote{To decode a particular test set, usually consisting of thousands of sentences, the phrase table can be drastically filtered by keeping only the phrase pairs applicable to the source sentences to translate. For the refinement steps of USMT, this filtering is impractical since we need to translate a very large number of sentences. In other words, it would still remain a large number of phrase pairs. Another alternative is to binarize the phrase table so that the system can load only applicable phrase pairs on-demand at decoding time. However, we did not consider it in our framework since the binarization is itself very costly to perform, and more importantly, the phrase table of each refinement step is used only once.} In SMT, decoding speed can be improved by reducing the size of the phrase table.
The phrase tables trained during the refinement steps are expected to be very noisy and very large since they are trained on noisy parallel data. Therefore, we assume that a large number of phrase pairs can be removed without sacrificing translation quality. On this assumption, we use the well-known algorithm for pruning phrase table \citep{johnson-EtAl:2007:EMNLP-CoNLL2007}, which has shown good performance in removing less reliable phrase pairs without any significant drop of the translation quality. This pruning can be done for each refinement step to reduce the phrase table size, and consequently to speed up the decoding.
Note that we cannot prune the induced phrase table used at iteration 0, since it was not learned from parallel data: we do not have co-occurrence statistics for the phrase pairs.

\section{UNMT as NMT trained exclusively on synthetic parallel data}

To make NMT able to learn how to translate from monolingual data only, previous work on UNMT \citep{artetxe2018unsupervisednmt,lample2018unsupervisednmt,DBLP:journals/corr/abs-1804-07755,P18-1005} proposed dedicated architectures, such as denoising autoencoders, shared latent representations, weight sharing, pre-trained sub-word embeddings, and adversarial training.

In this paper, we propose to train UNMT systems exclusively on synthetic parallel data, using existing frameworks for supervised NMT.
Specifically, we train the first UNMT system on synthetic parallel data generated by USMT through back-translating monolingual sentences in the target language, expecting that they are of a better quality than those generated by existing UNMT frameworks.

Our approach is significantly different from \citet{DBLP:journals/corr/abs-1804-07755}'s ``PBSMT+NMT'' configuration in the following two aspects.
First, while it uses synthetic parallel data generated by USMT only to further tune their UNMT system, ours uses it for initialization.
Second, they assumed certain level of relatedness between source and target languages, which is a prerequisite to jointly pre-train bilingual sub-word embeddings. Our approach does not make this assumption.

\begin{figure}[t]
    \centering
            \includegraphics[width=\linewidth]{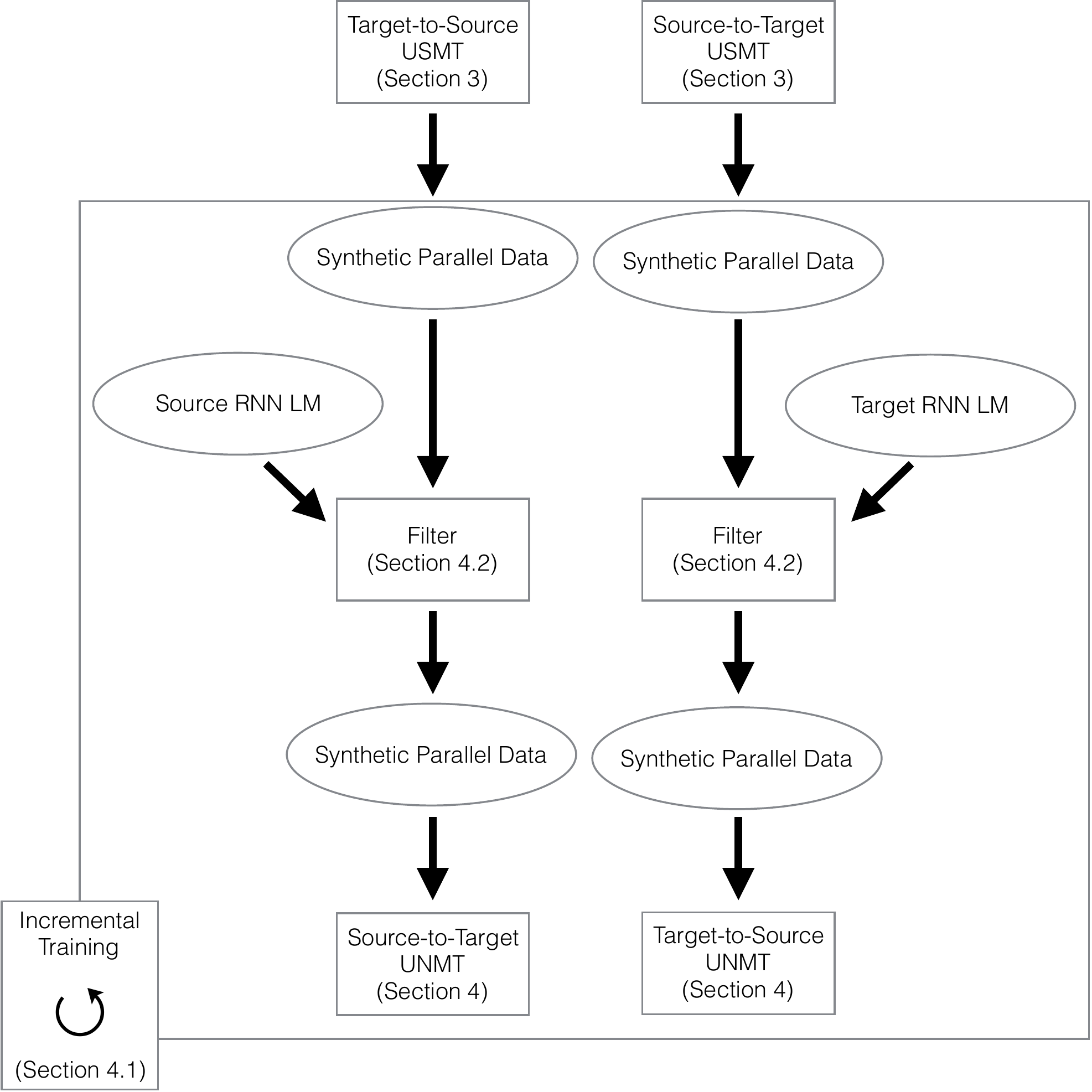}
    \caption{\label{fig:unmt} Our UNMT framework.}
\end{figure}

However, training an NMT system only on synthetic parallel data generated by USMT, as we proposed, will hardly make an UNMT system significantly better than USMT systems. To obtain better UNMT systems, we propose the following (see also Figure~\ref{fig:unmt}).
\begin{itemize}\itemsep=0mm
\item Section \ref{section:incr}: we propose an incremental training strategy for UNMT that gradually increases the quality and the quantity of synthetic parallel data.
\item Section \ref{section:filternmt}: we propose to filter the synthetic parallel data to remove before training the sentence pairs with the noisiest synthetic sentences, aiming at speeding up training and improving translation quality.
\end{itemize}

\subsection{Incremental training}
\label{section:incr}

To train UNMT, we first use the synthetic parallel data generated by the last refinement step of our USMT system. Since it has been shown that back-translated monolingual data significantly improves translation quality in NMT, as opposed to the refinement of our USMT (see \Sec{refinement}), we train source-to-target and target-to-source UNMT systems on synthetic parallel data respectively generated by a target-to-source and source-to-target USMT systems.

In contrast to supervised NMT where synthetic parallel data are used in combination with human-made parallel data, we can presumably use as much synthetic parallel data as possible, since seeing more and more fluent target sentences will be helpful to train a better decoder while we can assume that the quality of synthetic source side remains constant. In practice, generating a large quantity of synthetic parallel data is costly. Therefore, to train the first UNMT system, we use the same number, $N$, of synthetic sentence pairs generated by the final USMT system.

Since the source side of the synthetic parallel data is generated by USMT, it is expected to be of worse quality than those that state-of-the-art supervised NMT can generate. Therefore, we propose to refine UNMT through gradually increasing the quality and quantity of synthetic parallel data. First, we back-translate a new set of $N$ monolingual sentences using our UNMT systems at iteration 1 in order to generate new synthetic parallel data. Then, new UNMT systems at iteration 2 are trained from scratch on the $2N$ synthetic sentence pairs consisting of the new $N$ synthetic data and $N$ synthetic data generated by USMT. Note that we do not re-back-translate the monolingual data used at iteration 1 but keep them as they are for iteration 2 to reduce the computational cost. Similarly to the refinement steps of USMT, we can again perform this back-translation/re-training step for a pre-defined number of iterations to keep improving the quality of the source side of the synthetic data while increasing the number of new target sentences. At each iteration $i$, $(N\times i)$ synthetic sentence pairs are used for training.

This can be seen as an extension of \citet{hoang-EtAl:2018:WNMT20181}'s work, which performs a so-called iterative back-translation to improve NMT. The difference is that we introduce better synthetic parallel data, with new target sentences, at each iteration.

\subsection{Filtering of synthetic parallel data}
\label{section:filternmt}
Our UNMT system is trained on purely synthetic parallel data in which a large proportion of source sentences may be very noisy. We assume that removing the sentence pairs with the noisiest source sentences will improve translation quality.  Inevitably it also reduces the training time.

Each sentence pair in the synthetic parallel data is evaluated by the following normalized source language model score:
\begin{equation}
\label{eq:ppl}
ppl(S) = \frac{lm(S)}{len(S)+1}
\end{equation}
where $S$ is a (synthetic) source sentence, $lm(\cdot)$ the language model score, and $len(\cdot)$ a function returning the number of tokens in the sentence. We add 1 to the number of tokens to account for the special token used by NMT that marks the end of a sentence. This scoring function has a negligible computational cost, but has shown satisfying performances in our preliminary experiments.  While we do not limit the language model to be specific type, in our experiment, we use a recurrent neural network (RNN) language model trained on the entire source monolingual data. 

There are many ways to make use of the above score during NMT training. For instance, weighting the sentence pairs with this score during training is a possible alternative, and this idea is close to one used by \citet{ijcai2017-555} in their joint training framework for NMT. However, given that many of the source sentences would be noisy, we rather choose to discard potentially noisy pairs for training. It would also remove potentially useful target sentences, but we assume that the impact of this removal could be compensated at the succeeding iterations of UNMT, where we incrementally introduce new target sentences.

At each iteration $i$ of incremental training, we keep only the cleanest ($\alpha\times N\times i$) synthetic sentence pairs\footnote{We considered both the sentence pairs used to initialize UNMT and all the sentence pairs generated by each iteration of UNMT in the set of sentence pairs to filter.} selected according to the score computed by \Eq{ppl}, where $\alpha$ ($0<\alpha\le 1$) is the filtering ratio.\footnote{We used $\alpha=0.5$ in our experiments.} This aggressive filtering will speed up training while relying only on the most fluent sentence pairs.

\section{Experiments}
\label{section:exp}
In this section, we present experiments for evaluating our USMT and UNMT systems.
\subsection{Experimental settings}
\label{section:expset}
For these preliminary experiments, we chose the language pair English--German (en-de) and the evaluation task WMT16 (newstest2016) for both translation directions, following previous work \citep{artetxe2018unsupervised,DBLP:journals/corr/abs-1804-07755}. To train our USMT and UNMT, we used only monolingual data: English and German News Crawl corpora respectively containing around 238M and 237M sentences.\footnote{\url{http://www.statmt.org/wmt18/translation-task.html}}
All our data were tokenized and truecased with {\moses}'s tokenizer\footnote{We escaped special characters but did not use the option for ``aggressive'' tokenization.} and truecaser, respectively. The statistics for truecasing were learned from 10M sentences randomly sampled from the monolingual data.

For the phrase table induction, the source and target word embeddings were learned from the entire monolingual data with the default parameters of \texttt{fasttext} \citep{bojanowski2017enriching},\footnote{\url{https://fasttext.cc/}} except that we set to 200 the number of dimensions.\footnote{While \citet{artetxe2018unsupervised} and \citet{DBLP:journals/corr/abs-1804-07755} used 300 and 512 dimensions, respectively, we chose a smaller number of dimensions for faster computation, even though this might lead to lower quality.} For a reasonably fast computation, we retained only the embeddings for the 300k most frequent words. Word embeddings for two languages were then aligned in the same space using the \texttt{--unsupervised} option of \texttt{vecmap}.\footnote{\url{https://github.com/artetxem/vecmap}}  From the entire monolingual data, we also collected phrases of up to 6 tokens in each language using \texttt{word2phrase}.\footnote{\url{https://code.google.com/archive/p/word2vec/}} To maintain the experiments feasible and to make sure that we have a word embedding for all of the constituent words, we retained only 300k most frequent phrases made of words among the 300k most frequent words. We conserved the 300-best target phrases for each source phrase, according to \Eq{prob}, consequently resulting in the initial phrase table for USMT containing 90M (300k$\times$300) phrase pairs.

We used {\moses} and its default parameters to conduct experiments for USMT. The language models used by our USMT systems were 4-gram models trained with \texttt{LMPLZ} \citep{P13-2121} on the entire monolingual data.  In each refinement step, we trained a phrase table and a lexicalized reordering model on synthetic parallel data using \texttt{mgiza}.\footnote{\texttt{fast\_align} \citep{dyer-chahuneau-smith:2013:NAACL-HLT} is a significantly faster alternative for a similar performance on en-de \citep{W14-3309}. We used \texttt{mgiza} since it is integrated in {\moses}.} We compared USMT systems with and without supervised tuning. For supervised tuning, we used \texttt{kb-mira} \citep{N12-1047} and the WMT15 newstest (newstest2015). For the configurations without tuning, we used {\moses}'s default weights as in previous work.

For UNMT, we used the Transformer \citep{NIPS2017_7181} model implemented in {\marian} \citep{P18-4020}\footnote{\url{https://marian-nmt.github.io/}, version 1.6.} with the hyper-parameters proposed by \citet{NIPS2017_7181}.\footnote{Considering the computational cost of our approach for UNMT, we did not experiment with the ``big'' version of the Transformer model while it would probably have resulted in a better translation quality.} We reduced the vocabulary size by using byte-pair-encoding (BPE) with 8k symbols jointly learned for English and German from 10M sentences sampled from the monolingual data. BPE was then applied to the entire source and target monolingual data.\footnote{We did not use BPE for USMT.} We used the same BPE vocabulary throughout our UNMT experiments.\footnote{Re-training BPE at each iteration of UNMT on synthetic data did not improve the translation quality in our preliminary experiments.} We validated our model during UNMT training as proposed by \citet{DBLP:journals/corr/abs-1804-07755}: we did a supervised validation using 100 human-made sentence pairs randomly extracted from newstest2015. We consistently used the same validation set throughout our UNMT experiments. To filter the synthetic parallel sentences (see Section \ref{section:filternmt}), we used an RNN language model trained on the entire monolingual data, without BPE, with a vocabulary size of 100k.\footnote{We used also {\marian} to train the RNN language models.}
\begin{table*}[t]
\centering
\begin{tabular}{lcccc}
\toprule
 System & USMT Tuning & de$\rightarrow$en & en$\rightarrow$de & \# \\ 
\midrule
 \citet{DBLP:journals/corr/abs-1804-07755} USMT & No  & 22.7* & 17.8* & 1\\
 \citet{artetxe2018unsupervised}  USMT    & back-translation & 23.1* & 18.2* & 2 \\
\midrule
\multirow{2}{*}{USMT (this work) w/ back-translation}  & Supervised  & 20.5  & 17.0 & 3\\
 & No   & 19.5  & 15.0  & 4\\ 
 \midrule
\multirow{2}{*}{USMT (this work)}& Supervised  &  22.1 & 17.4 & 5 \\
 & No  & 20.2  &15.5 & 6 \\
 \midrule
\citet{DBLP:journals/corr/abs-1804-07755} UNMT & No & 21.0* & 17.2* & 7\\
\midrule
\citet{DBLP:journals/corr/abs-1804-07755} USMT+UNMT & No  & 25.2* & 20.2* & 8\\
\midrule 
  \multirow{2}{*}{UNMT (this work) w/o filtering } & Supervised   & 28.2  & 21.3 & 9 \\
 & No & 27.0 &19.6  & 10 \\
 \midrule
\multirow{2}{*}{UNMT (this work)} & Supervised  & 28.8  & 21.6 & 11 \\
 & No  & 26.7  & 20.0 & 12\\
\midrule
\midrule
Supervised NMT (1.4M sent. pairs) & Supervised  & 32.5  & 29.9 & 13 \\
Supervised NMT (2.8M sent. pairs) & Supervised  & 33.8  & 31.6 & 14 \\
Supervised NMT (5.6M sent. pairs) & Supervised  & 34.9  & 32.3 & 15 \\
\bottomrule
\end{tabular}
\caption{\label{tab:results} Results of our USMT and UNMT systems (denoted ``this work'') evaluated with BLEU for the WMT16 German--English news translation task.  We present results for USMT with back-translation (\#3 and \#4) and forward translation (\#5 and \#6) during the refinement steps. Results for UNMT are presented without (\#9 and \#10) and with (\#11 and \#12) filtering of synthetic parallel data. ``*'' indicates the scores shown in the original paper for indicative purpose only, since they are tokenized BLEU scores and thus not directly comparable with our results.}
\end{table*}

For each of USMT and UNMT, we performed 4 refinement iterations. USMT has one more system in the beginning, which exploits an induced phrase table. At each iteration, we sampled new 3M monolingual sentences: i.e., $N=3000000$.\footnote{\citet{artetxe2018unsupervised} and \citet{DBLP:journals/corr/abs-1804-07755} respectively sampled 2M and 5M monolingual sentences.}

For reference, we also trained supervised NMT with {\marian} on 5.6M, 2.8M, and 1.4M human-made parallel sentences provided by the WMT18 conference for the German--English news translation task.\footnote{We did not use the ParaCrawl corpus.}

We evaluated our systems with detokenized and detruecased BLEU-cased \citep{papineni-EtAl:2002:ACL}. Note that our results should not be directly compared with the tokenized BLEU scores reported in \citet{artetxe2018unsupervised} and \citet{DBLP:journals/corr/abs-1804-07755}. 

\begin{figure*}[t]
    \centering
     \begin{subfigure}[b]{0.48\textwidth}
        \includegraphics[width=\textwidth]{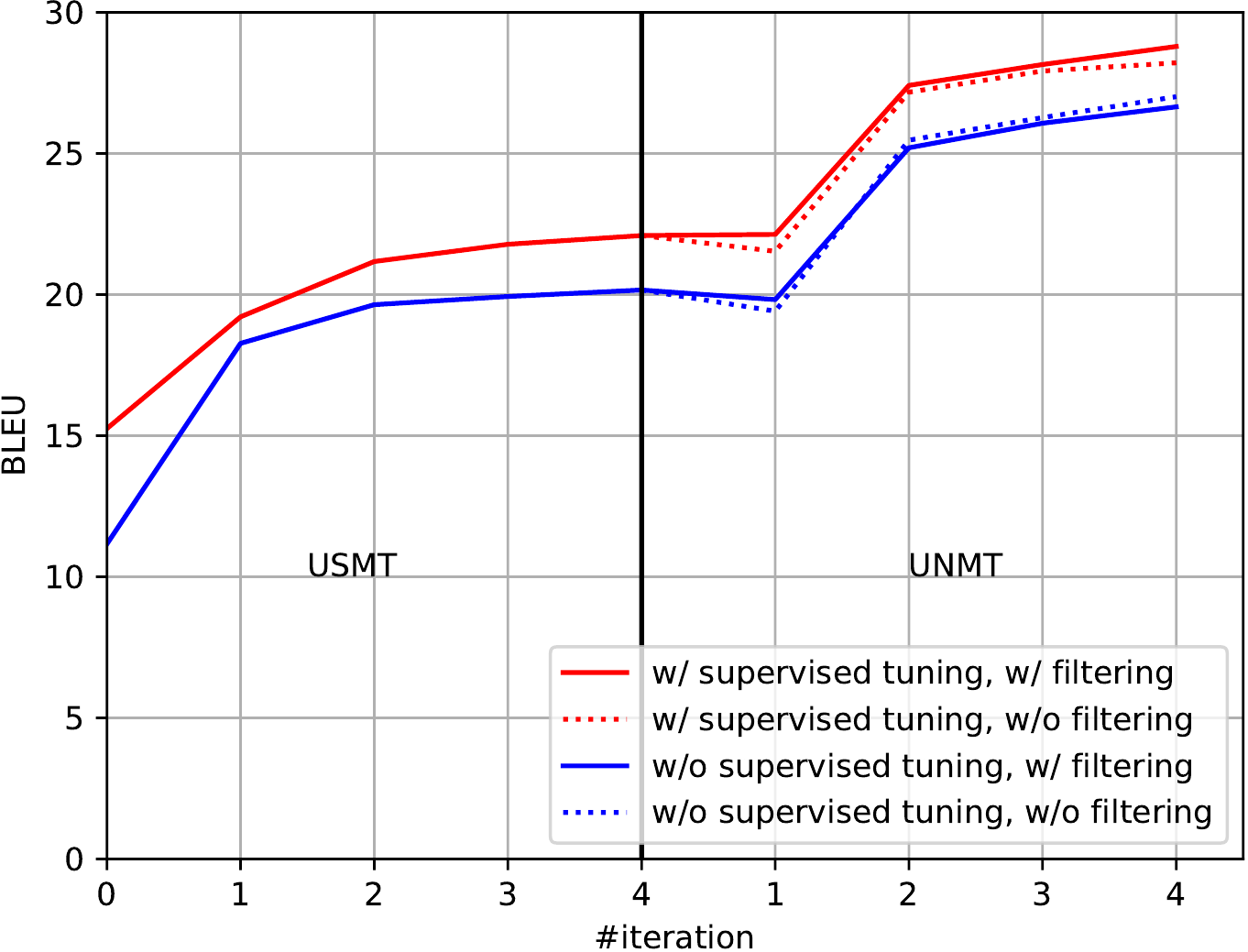}
        \caption{de$\rightarrow$en}
        \label{fig:deenlc}
    \end{subfigure}
    \begin{subfigure}[b]{0.48\textwidth}
        \includegraphics[width=\textwidth]{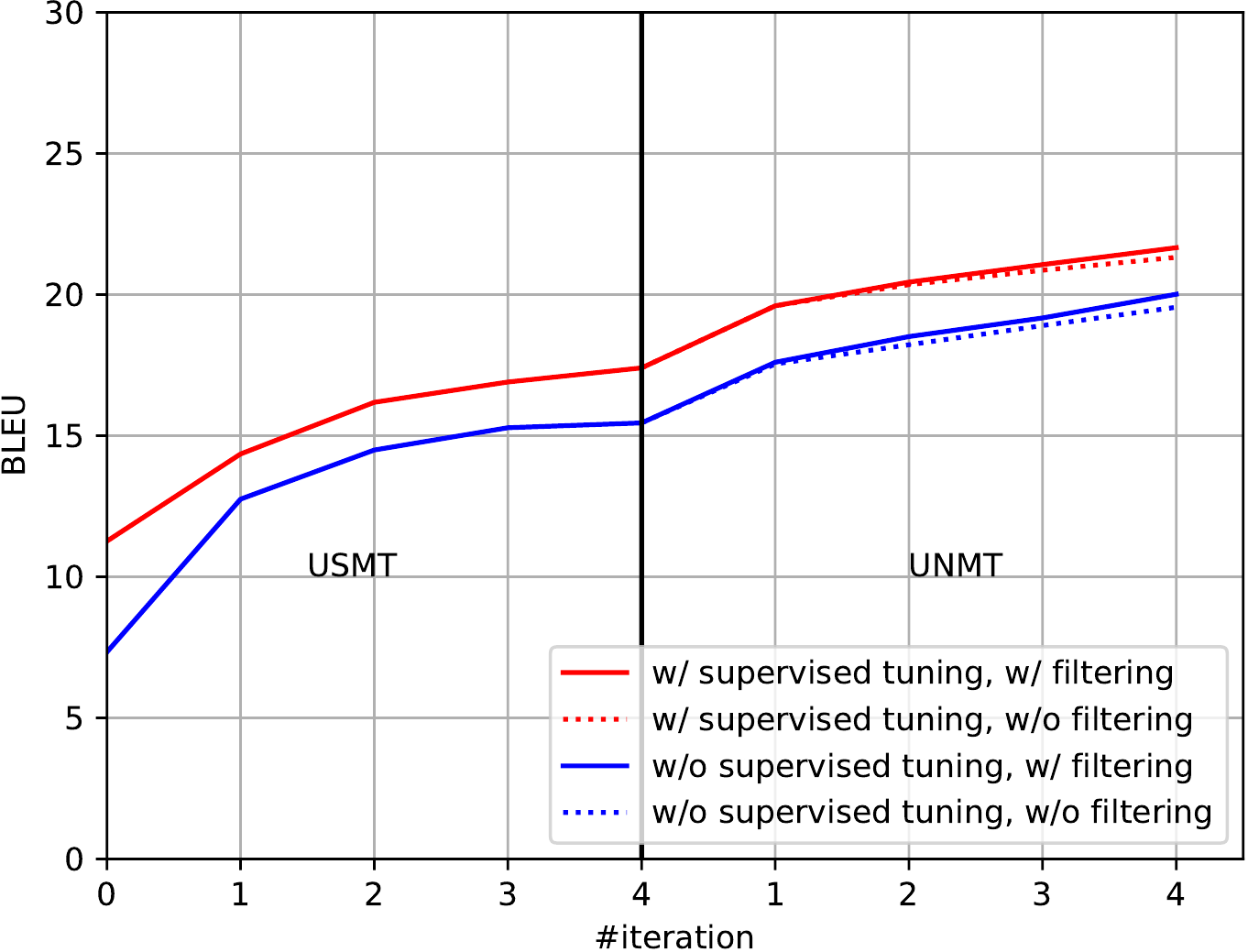}
        \caption{en$\rightarrow$de}
        \label{fig:endelc}
    \end{subfigure}
    \caption{\label{fig:lc}Learning curves of our USMT (\#5 and \#6) and UNMT (\#9, \#10, \#11, and \#12) systems presented in Section \ref{section:exp}.}
\end{figure*}

\subsection{Results}

Our results for USMT and UNMT are presented in Table \ref{tab:results}. 

We can first observe that supervised tuning for USMT improves translation quality, with 2.0 BLEU points of improvements, for instance between systems \#5 and \#6. Another interesting observation is that this improvement is carried on until the final iteration of UNMT (\#11 and \#12). These results show the importance of development data for tuning that could be created at a reasonable cost (see Section \ref{section:tuning}).

Our USMT systems benefited more from forward translation (\#5 and \#6) than back-translation (\#3 and \#4) during the refinement steps, with an improvement of 1.6 and 0.4 BLEU points for de$\rightarrow$en and en$\rightarrow$de (with supervised tuning), respectively. Pruning the phrase table (see Section \ref{section:pruning}) did not hurt translation quality but removed around 93\% of the phrase pairs in the phrase tables for each refinement step. Nonetheless, our USMT systems seem to significantly underperform the state-of-the-art USMT proposed by \citet{DBLP:journals/corr/abs-1804-07755} (\#1) and \citet{artetxe2018unsupervised} (\#2).  This is potentially the consequence of the following: we used much lower dimensions for our word embeddings and much less phrases (300k source and target phrases), than in \citet{artetxe2018unsupervised} (1M source and target phrases). In our future work, we will investigate whether their parameters improve the performance of our USMT systems.

While our USMT systems do not seem to outperform previous work, we can observe that the synthetic parallel data that they generated are of sufficient quality to initialize our UNMT. Incremental training improved significantly translation quality. To the best of our knowledge, we report the best results of unsupervised MT for this task which is, for de$\rightarrow$en, only 3.7 BLEU points lower (\#11) than a supervised NMT system trained on 1.4M parallel sentences (\#13).\footnote{A fair supervised NMT baseline should also use, in addition to human-made parallel sentences, back-translated data for training.} Our best UNMT systems (\#11 and \#12) significantly outperformed our USMT systems (\#5 and \#6) by more than 6.0 BLEU points, for de$\rightarrow$en.
Filtering the synthetic parallel sentences at each iteration significantly improved the training speed\footnote{For instance, for the last iteration of UNMT for de$\rightarrow$en, the training using 4 GPUs consumed 30 hours with filtering while it took 52 hours without filtering.} for a comparable or better translation quality for both translation directions. The results confirm the importance of filtering the very noisy synthetic source sentences generated by back-translation.

\subsection{Learning curves}

In this section, we present the evolution of the translation quality during training of USMT and UNMT.

The learning curves of our systems, for the same experiments presented in Section \ref{section:expset}, are given in Figures \ref{fig:deenlc} and \ref{fig:endelc} for de$\rightarrow$en and en$\rightarrow$de, respectively.
Iteration 0 of our USMT, using an induced phrase table, performed very poorly; for instance systems without supervised tuning (leftmost points of blue lines) achieved only 11.2 and 7.3 absolute BLEU points for de$\rightarrow$en and en$\rightarrow$de, respectively.  Iterations 1 and 2 of USMT were very effective and covered most of the improvements between iteration 0 and iteration 4. After 4 iterations, we observed improvements of 9.0 and 8.1 BLEU points for de$\rightarrow$en and en$\rightarrow$de, respectively.

The learning curves of UNMT were very different for the two translation directions.
The first iteration of UNMT, trained on the synthetic parallel data generated by USMT, performed slightly lower than USMT for de$\rightarrow$en while for en$\rightarrow$de we observed around 2.0 BLEU points of improvements. This confirms the ability of NMT in generating significantly better sentences than SMT for morphologically-rich target languages \citep{bentivogli-EtAl:2016:EMNLP2016}. Then, the second iteration of UNMT improved the translation quality significantly for de$\rightarrow$en, but much more moderately for en$\rightarrow$de. For instance, in the configuration without supervised tuning and with language model filtering (blue solid lines), we observed 5.4 and 0.9 BLEU points of improvements for de$\rightarrow$en and en$\rightarrow$de, respectively. Succeeding iterations continued to improve translation quality but more moderately.

For both translation directions, the learning curves highlighted that improving the synthetic parallel data generated by USMT, and used to initialize UNMT, is critical to improve UNMT: synthetic parallel data generated with tuned USMT were consistently more useful for UNMT than the synthetic parallel data of lower quality generated by USMT without tuning.

\section{Conclusion an future work}
We proposed a new approach for UNMT that can be straightforwardly exploited with well-established architectures and frameworks used for supervised NMT without any modifications. It only assumes for initialization the availability of synthetic parallel data that can be, for instance, easily generated by USMT. We showed that improving the quality of the synthetic parallel data used for initialization is crucial to improve UNMT. We obtained with our approach a new state-of-the-art performance for unsupervised MT on the WMT16 German--English news translation task.

For future work, we will extend our experiments to cover many more language pairs, including distant language pairs for which we expect that our approach will perform better than previous work that assumes the relatedness between source and target languages. We will also analyze the impact of using synthetic parallel data of a much better quality to initialize UNMT. Moreover, we would like to investigate the use of much noisier and not comparable source and target monolingual corpora to train USMT and UNMT, since we consider it as a more realistic scenario when dealing with truly low-resource languages. We will also study our approach in the semi-supervised scenario where we assume the availability of some human-made bilingual sentence pairs for training.
 
\bibliography{tacl2018}
\bibliographystyle{acl_natbib}

\end{document}